%% file: main.tex
\documentclass[10pt,twocolumn,letterpaper]{article}

\usepackage{cvpr}              

\input{preamble}

\definecolor{cvprblue}{rgb}{0.21,0.49,0.74}
\usepackage[pagebackref,breaklinks,colorlinks,citecolor=cvprblue]{hyperref}

\def\paperID{77} 
\def\confName{3DV\xspace}
\def\confYear{2025\xspace}

\title{FastGrasp: Efficient Grasp Synthesis with Diffusion}

\author{
Xiaofei Wu$^1$, Tao Liu$^1$, Caoji Li$^1$, Yuexin Ma$^1$, Yujiao Shi$^{1*}$, Xuming He$^{1,2*}$\\
$^1$ShanghaiTech University, Shanghai, China \\
\textsuperscript{2}Shanghai Engineering Research Center of Intelligent Vision and Imaging \\
{\tt\small{\{wuxf2023, liutao2023, licj, mayuexin, shiyj2, hexm\}@shanghaitech.edu.cn}}
}

\begin{document}
\twocolumn[{
\renewcommand\twocolumn[1][]{#1}%
\maketitle

\begin{center}
\setlength{\abovecaptionskip}{0pt}
  \setlength{\belowcaptionskip}{0pt}
    \vspace{-3ex}
    \captionsetup{type=figure}
    \includegraphics[width=\textwidth]{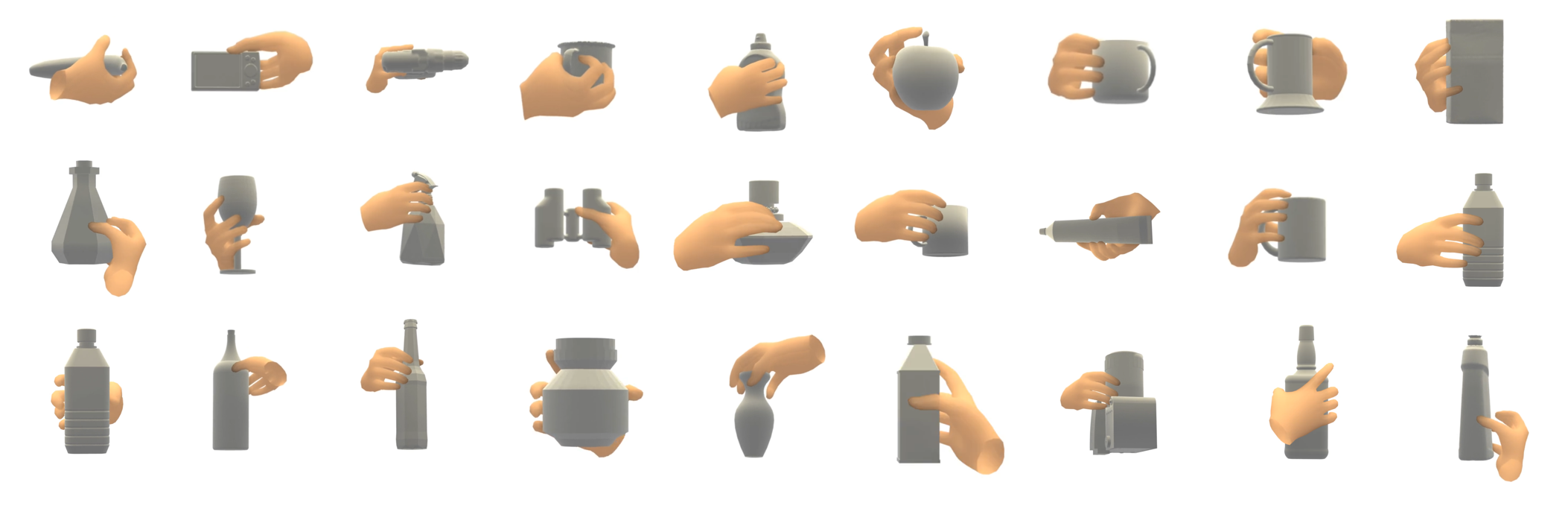}
    \vspace{-3ex}
    \caption{\textbf{FastGrasp} provides extensive realistic grasping of dexterous hands synchronized with human poses.}
    \label{fig:teaser}
\end{center}
}]

\renewcommand{\thefootnote}{*}
\footnotetext{Corresponding authors.}
\input{sec/0_abstract}    
\input{sec/1_intro}
\input{sec/2_relat}
\input{sec/3_approach}

\input{sec/4_exp}

\input{sec/5_conclusion}
{
    \small
    \bibliographystyle{ieeenat_fullname}
    \bibliography{main}
}

\input{sec/X_suppl}

\end{document}

%% file: preamble.tex
\usepackage[dvipsnames]{xcolor}


%% file: sec/0_abstract.tex
\begin{abstract}
\vspace{-1mm}
Effectively modeling the interaction between human hands and objects is challenging due to the complex physical constraints and the requirement for high generation efficiency in applications. 
Prior approaches often employ computationally intensive two-stage approaches, which first generate an intermediate representation, such as contact maps, followed by an iterative optimization procedure that updates hand meshes to capture the hand-object relation. 
However, due to the high computation complexity during the optimization stage, such strategies often suffer from low efficiency in inference. 
To address this limitation, this work introduces a novel diffusion-model-based approach 
that generates the grasping pose in a one-stage manner. This allows us to significantly improve generation speed and the diversity of generated hand poses. 
In particular, we develop a Latent Diffusion Model with an Adaptation Module for object-conditioned hand pose generation and a contact-aware loss to enforce the physical constraints between hands and objects.     
Extensive experiments demonstrate that our method achieves faster inference, higher diversity, and superior pose quality than state-of-the-art approaches. 
Code is available at \href{https://github.com/wuxiaofei01/FastGrasp}{https://github.com/wuxiaofei01/FastGrasp.}



\end{abstract}

%% file: sec/1_intro.tex
\section{Introduction}
\label{sec:intro}
The problem of modeling hand-object interactions~\cite{intro-hand-object-interaction-1,intro-hand-object-interaction-2,Doosti2020HOPENetAG,Liu2021SemiSupervised3H,Chen2023gSDFGS,Motion-Generation-for-Hand-Object-Interaction} has attracted increasing research interest recently, with important applications in virtual reality~\cite{Costabile2005HumancomputerI}, human-computer interaction~\cite{Hll2018EfficientPI,Farulla2016VisionBasedPE}, and imitation learning in robotics. 
A key task in hand-object interaction modeling is to predict 
various ways a human hand can grasp a given object. 
Unlike robot grasping with parallel jaw grippers, the task of predicting human grasps is particularly challenging due to two reasons: First, human hands have more degrees of freedom, resulting in more intricate contact patterns; Moreover, the generated grasp must be not only physically plausible but also appear natural, reflecting the typical ways that humans handle objects. 

Previous methods for synthesizing human grasping postures often rely on a two-stage process~\cite{gf,halo,liu2023contactgen,tta}. Such a process typically first uses a generative model, e.g., Conditional Variational AutoEncoder (CVAE)~\cite{cvae}, to generate a series of intermediate representations, including contact maps~\cite{tta} and/or parts maps~\cite{liu2023contactgen}, based on the point cloud representation of interacting objects. 
The second stage then uses those intermediate representations to estimate the hand parameters, aiming to produce a natural and physically plausible hand pose. To achieve this, most methods formulate the estimation as an optimization problem and adopt an iterative procedure to search the target hand pose.
Despite their promising results, such two-stage methods often suffer from two drawbacks: First, the iterative optimization procedures are computationally intensive, leading to a low inference efficiency and time-consuming generation; Second,  the quality of generated hand poses highly relies on the intermediate representations from the first stage and prone to accumulated errors. 

To address those limitations, we propose an efficient one-stage generation method, named \textit{FastGrasp}, to directly generate grasping poses without producing intermediate representations like contact maps, while maintaining the diversity of generated poses. 
To achieve this, we leverage the latent diffusion model framework~\cite{Preechakul2021DiffusionAT} to learn a contact-aware representation for hand poses in a latent space and a diffusion-based generation process, capable of better encoding the physical constraints and capturing the object-conditioned hand-pose distribution. 

Specifically, \textit{FastGrasp} first learns a low-dimensional latent representation of hand pose parameters based on an AutoEncoder (AE) network. It then encodes the object with a Point-Net and builds a diffusion model in the latent space conditioned on the object representation. 
Subsequently, to incorporate the physical constraints on hand-pose interaction, \textit{FastGrasp} introduces an adaptation module, which refines the diffusion-generated latent representation based on the object contact information. Finally, the contact-aware hand-pose presentation is decoded into the MANO~\cite{mano} parameters of the grasping hand pose with the AE decoder.

We validate our approach through extensive experiments on three hand-object interaction benchmarks: HO-3D~\cite{ho3d}, OakInk~\cite{YangCVPR2022OakInk}, and Grab~\cite{grab}.
Experimental results demonstrate that our method achieves low latency in inference and generates higher-quality grasping poses with more plausible physical interactions and higher diversity than recent state-of-the-art approaches. 

In summary, our contributions are as follows:
\begin{itemize}
\item We introduce FastGrasp, a diffusion-based one-stage model for generating grasping hand pose without requiring expensive iterative optimization. 
\item  We propose an adaptation module to effectively incorporate physical constraints into a latent hand representation. 
\item  Our approach achieves fast inference and outperforms previous state-of-the-art methods on a range of metrics.
\end{itemize}

%% file: sec/2_relat.tex
\section{Related Work}
\subsection{Hand-object Interaction}
Generating whole-body interactions, such as approaching and manipulating static~\cite{Kulkarni2023NIFTYNO,wu2022saga} and dynamic objects~\cite{Ghosh2022IMoSIF}, is a growing topic. The task of synthesizing humans interacting with dynamic objects is explored using first-person vision~\cite{Liu2019ForecastingHO} in skeleton-based datasets. 
However, numerous studies begin to explore hand-object interactions across diverse settings~\cite{Brahmbhatt2019ContactDBAA,Brahmbhatt2020ContactPoseAD,Chao2021DexYCBAB,Liu2021SemiSupervised3H}. Most current efforts focus on synthesizing these interactions in the domains of computer graphics~\cite{Li2007DataDrivenGS,Pollard2005PhysicallyBG,Zhang2021ManipNet}, computer vision~\cite{Grady2021ContactOptOC,Jiang2021HandObjectCC,Kry2005InteractionCA,Li2023TaskOrientedHI,Ye2023AffordanceDS,Zheng2023CAMSCM,Zhou2022TOCHSO}, and robotics~\cite{Brahmbhatt2019ContactGraspFM,Detry2010RefiningGA,Hsiao2006ImitationLO}.
To perform hand-object pose estimation, Tekin \etal.~\cite{Tekin2019HOUE} proposes a 3D detection framework that predicts hand-object poses using two output grids without explicitly modeling their interaction. In contrast, Hasson \etal.~\cite{Hasson2019LearningJR} utilize hand-centric physical constraints to model hand-object interactions and prevent penetration. Recently, research shifts towards generating plausible hand grasps for objects, with significant contributions including:~\cite{Corona2020GanHandPH,grab}. GanHand~\cite{Corona2020GanHandPH} generates grasps suitable for each object in a given RGB image by predicting a grasp type from grasp taxonomy and its initial orientation, then optimizing for better contact with the object. GrabNet ~\cite{grab} represents 3D objects using Basis Point Set to generate MANO~\cite{mano} parameters. The predicted hand is refined using an additional model to enhance contact accuracy. Our diffusion-model-based pipeline directly generates the grasping pose for a given object point cloud, eliminating the need for additional models.

\subsection{Grasp Synthesis}
Grasp synthesis receives extensive attention across robotic hand manipulation, animation, digital human synthesis, and physical motion control~\cite{wu2022saga,wxf}. In this work, we focus on realistic human grasp synthesis~\cite{gf,grab,halo,liu2023contactgen,tta}, aiming to generate authentic human grasps for diverse objects. The key challenge is achieving physical plausibility and generation efficiency. Most existing approaches employ CVAE to generate hand MANO parameters~\cite{grab,tta,YangCVPR2022OakInk} or hand joints~\cite{halo}. Liu \etal.~\cite{liu2023contactgen} propose learning intermediate representations followed by iterative optimization in two stages. This method weakens the spatial information of objects, causing intersection penetration and displacement, and requires significant time for optimization in the second stage. In contrast, we develop an one-stage generation model that supervises the spatial geometry of objects and adaptively learns the physical constraints of hand-object interaction. Such model architecture effectively accelerates generation speed and reduces hand-object penetration volume.
\begin{figure*}[ht]
    \setlength{\abovecaptionskip}{0pt}
    \setlength{\belowcaptionskip}{0pt}

    \centering
    \includegraphics[width=\textwidth]{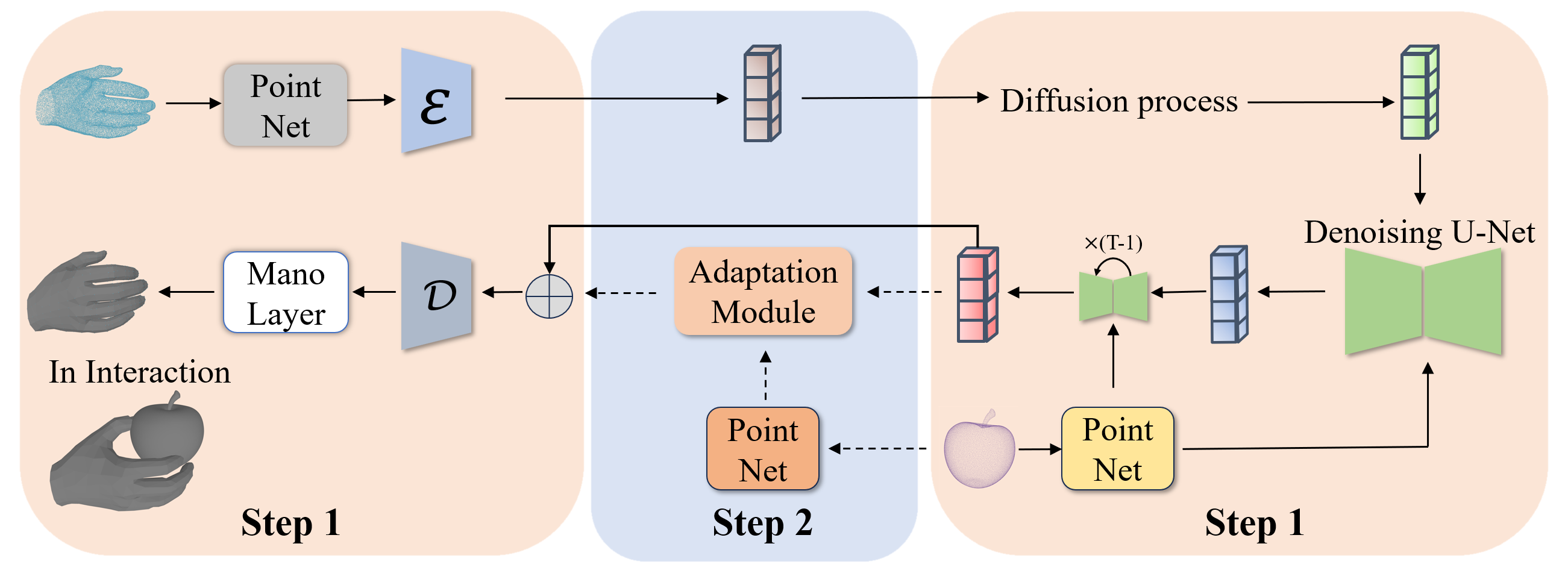}
    \caption{\textbf{Model training architecture.} 
    We divide the training process into two parts. In the first part, we use a latent diffusion model to generate grasping poses from object point clouds. However, the diffusion model struggles to directly learn the physical constraints between the hand and object, leading to issues such as penetration and displacement. To address this, the second part involves training an Adaptation Module to refine the grasping gestures by aligning them with the physical constraints of hand-object interactions, resulting in more natural and feasible poses. In training stage one, only the solid arrow path is utilized. In stage two, both the solid and dotted arrow paths are used.}
    \label{fig:train-pipeline}
\end{figure*}
\subsection{Denoising Diffusion Probabilistic Models}
Denoising diffusion models~\cite{ho2020denoising,SohlDickstein2015DeepUL,Liu2022CompositionalVG,Poole2022DreamFusionTU,Watson2022NovelVS} utilize a stochastic diffusion process that incrementally introduces noise into a sample from the data distribution, adhering to thermodynamic principles. They then generate denoised samples through a reverse iterative procedure. However, directly training DDPMs on high-resolution point clouds and sampling from them is computationally intensive. Latent diffusion models address this issue by encoding high-resolution images into a low-dimensional latent space~\cite{lyu2023controllable,Kwon2022DiffusionMA,Preechakul2021DiffusionAT} before training DDPMs. Our approach follows this paradigm: we first train an autoencoder in the data space, and then train a DDPM using the encoded samples. Additionally, we designe an Adaptation Module(AM) to adjust the input to the decoder, incorporating hand-object physical constraints into the diffusion model.



%% file: sec/3_approach.tex
\section{Fast Grasping Hand Pose Generation}

\subsection{Method Overview}

Given an object, usually represented by a point cloud, our purpose is to generate a human hand pose for grasping this object. 
The generated grasping hand pose should be natural and physically correct, 
securely holding the object in a physically plausible manner.
Unlike the existing methods that usually adopt a two-stage design with high computation cost,  
we propose \textit{FastGrasp}, a fast grasping hand pose generation pipeline without estimating intermediate representations and iterative optimizations. 

\textit{FastGrasp} is a one-stage generation framework consisting of two main modules for generating the grasping hand pose. The first module is based on a latent diffusion model to preserve the diversity of hand poses when intermediate representations like contact maps are absent. Given the latent hand representation generated from the diffusion model, we introduce an adaptation module to enforce the physical constraints of hand-object interaction. This design allows the model to directly learn the spatial relationship between the hand and object point clouds without iterative optimization, resulting in a fast generation of high-quality hand poses.

To learn the entire model, we adopt a simple yet effective two-step training strategy. The first step trains the latent diffusion model, which generates an initial representation of the hand poses. Next, we train the adaptation module to refine the hand representation to strengthen the physical constraints of the hand-object interaction. After training, our generation requires only one pass of network inference, thus significantly accelerating grasping hand generation. 

Below we will first introduce the latent diffusion model module in Sec.~\ref{latent-diffusion}, followed by the adaptation module in Sec.~\ref{distribution-adapt-layer}. Finally, the model inference pipeline will be detailed in Sec.~\ref{inference}.

\subsection{Latent Diffusion Model for Hand Pose}
\label{latent-diffusion}

\paragraph{Latent Hand Representation.}
To build our Latent Diffusion model~\cite{Kwon2022DiffusionMA} for hand pose, we first train an auto-encoder that maps the input hand representation to a latent space. 
This allows us to reduce the data dimensionality for the diffusion process and improves the modeling efficiency. 
In contrast to the original latent diffusion model, where the input and output are exactly the same, we employ an asymmetric design in the auto-encoder for the subsequent conditional generation process. 

Specifically, the input to our auto-encoder is the vertices of the hand mesh, $h_v \in \mathbb{R}^{778 \times 3}$, which is first processed by a PointNet~\cite{Pointnet} and then fed into the encoder block. This design maintains the spatial shape information of the input hand in feature extraction, which can be more easily integrated with the object representation in the later stage. The obtained latent vector is converted to MANO~\cite{mano} parameters representation $h_p \in \mathbb{R}^{61}$ instead of the vertices by the decoder block. The MANO parameters have far less freedom than those of vertices, thus improving the regularization in learning the decoder. 
The hand mesh vertices $h_m \in \mathbb{R}^{778 \times 3}$ is finally reconstructed from $h_p$ by a differentiable MANO layer~\cite{mano}. 


The training objective of the AutoEncoder combines a hybrid reconstruction loss and a set of physical constraints. The reconstruction loss measures the difference between the reconstructed hand mesh and the ground truth, which includes two terms:
\begin{align}
   \mathcal{L}_{recon} & = \lambda_1 \mathcal{L}_{param} + \lambda_2\mathcal{L}_{mesh}\\
   \mathcal{L}_{param} &= \text{MSE}(h_p, h_p^{\text{gt}})\label{eq:LOSS-param} \\
   \mathcal{L}_{mesh} &= \text{Chamfer-Dis}(h_m, h_m^{\text{gt}}) \label{eq:LOSS-mesh}
\end{align}
where $\mathcal{L}_{param}$ indicates mean squared error loss between predicted $h_p$ and GT hand MANO parameters $h_p^{gt}$, $\mathcal{L}_{mesh}$ measures chamfer distance between the predicted hand vertices $h_m$ and the GT hand vertices $h_m^{gt}$. $\lambda_{1}$ and $\lambda_{2}$ are the weight balancing coefficients.

To learn a hand representation that adheres to physical constraints, we also employ the following three loss functions from~\cite{tta}:
\begin{equation}
\mathcal{L}_{consist} = \text{Consist}(h_m, h_m^{gt}, o_m)
\label{eq:consist}
\end{equation}
\begin{equation}
\mathcal{L}_{cmap} = \text{Contact}(h_m, o_m)
\label{eq:cmap}
\end{equation}
\begin{equation}
\mathcal{L}_{penetr} = \text{Penetra}(h_m, o_m)
\label{eq:penetr}
\end{equation}
where $o_m$ denotes the object mesh that we aim to grasp, $\mathcal{L}_{consist}$ aims to make the contact region of the predicted hand mesh on the object consistent with that of the GT hand mesh on the object. 
$\mathcal{L}_{cmap}$ ensures that the hand mesh generated by the model maintains contact with the object. 
$\mathcal{L}_{penetr}$ prevents the hand mesh and objects from penetrating the physical volume. 
We refer the reader to the Supplementary for details of those loss functions. 

Our total loss function for training the auto-encoder (the left part in Fig.~\ref{fig:train-pipeline}) can be written as:
\begin{align}
    \mathcal{L} = & \mathcal{L}_{recon} + \lambda_{3}\mathcal{L}_{cmap} + \lambda_{4}\mathcal{L}_{penetr} + \lambda_{5}\mathcal{L}_{consist}
  \label{eq:Total-loss}
\end{align}
where $\lambda_{3},  \lambda_{4},  \lambda_{5}$ are weight parameters for balancing the physical constraint loss terms.
By integrating physical and reconstruction losses, our model is able to learn the hand mesh and the physical constraints involved in the interaction between the hand and the object. This approach ensures that our auto-encoder effectively encodes the hand vertices and maintains the physical plausibility of the generated mesh.
\begin{figure*}[ht]
    \centering
    \setlength{\abovecaptionskip}{0pt}
    \setlength{\belowcaptionskip}{0pt}
    \includegraphics[width=\textwidth]{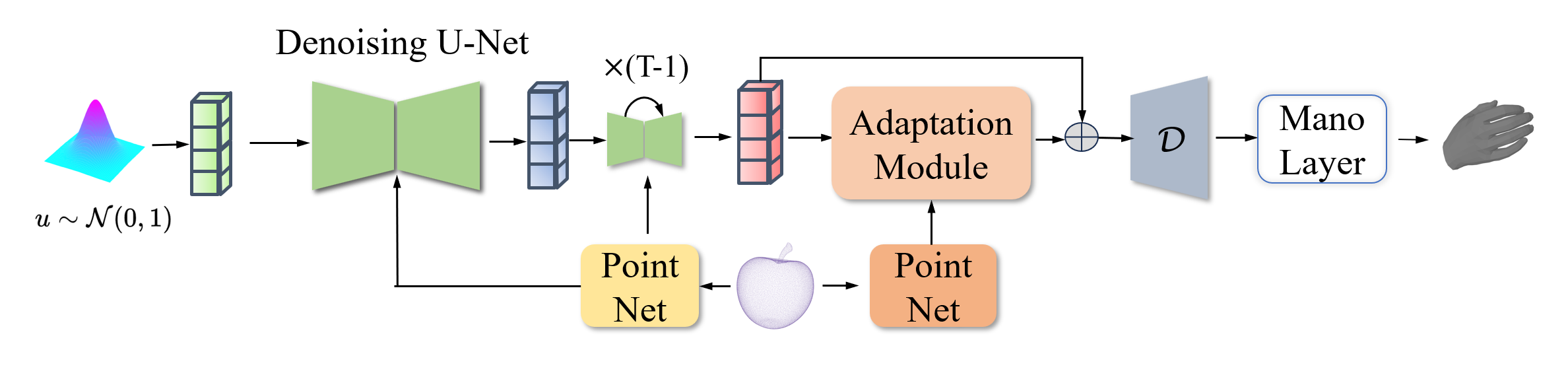}
    \caption{\textbf{Model inference architecture.} We start by inputting Gaussian noise and the object’s point cloud into the model. The diffusion model then generates hand representations in latent space. The  Adaptation Module refines these representations, which are then decoded into MANO parameters. Finally, we construct the hand mesh using the MANO layer.}
    \label{fig:inference-pipeline}
\end{figure*}

\paragraph{Diffusion Model for Hand Representations.} 
We adopt a diffusion model to learn the distribution of the latent hand representation produced by the auto-encoder. The model gradually denoises a normally distributed random variable, which corresponds to learning the reverse process of a fixed Markov Chain~\cite{ho2020denoising,rombach2021highresolution}. Here we train a denoising U-Net to predict the added noises in the diffusion process, as shown in the right part of Fig. \ref{fig:train-pipeline}. 

Specifically, the input of the diffusion model consists of three parts: $z_0, o_p \in R^{N_o \times 3}$ and $t$. $z_0$ be the feature output of the encoder $\mathcal{E}$ when the input is $h_v$. The input object point cloud $o_p$, is used as the conditioning information for our diffusion model. It is transformed into an embedding using PointNet~\cite{Pointnet}, facilitating controllable generation. $t$ denotes the time step in the diffusion model training process. 
The loss function for training the diffusion network can be written as:
\begin{equation}
  L_{LDM}:=\mathbb{E}_{\mathcal{E}(h_v),\epsilon\thicksim\mathcal{N}(0,1),t}\Big[\|\epsilon-\epsilon_\theta(z_0^t,P(o_p),t)\|_2^2\Big]
  \label{eq:diff1}
\end{equation}
where $\epsilon_\theta(z_0^t,P(o_p),t)$ denotes the conditional denoising U-Net used for training, where $t$ ranges from 1 to $T$, the input $z_0^t$  is the  $z_0$  mixed with $\epsilon$, the $P$ denotes the PointNet~\cite{Pointnet}. Through training, the diffusion model learns to reconstruct the hand mesh from Gaussian noise by denoising and decoding.

\subsection{Physical Constraints Alignment}
\label{distribution-adapt-layer}
During the training of the diffusion model, directly incorporating physical loss and reconstruction loss lead to oscillations and hampers convergence. We attribute this issue to the diffusion model's difficulty in simultaneously learning the distribution of the $\mathcal{E}$ output and capturing the physical constraints between the hand and the object. Therefore, the generated hand mesh and object may exhibit significant physical penetration and displacement. To address this problem, we decompose the entire training process into a two-step optimization approach. This method not only simplifies the model’s training complexity but also helps better capture the physical constraint relationship between the hand and the object.

Specifically, after training the diffusion model, we aim to adjust the physical constraints of hand-object interactions. To retain the knowledge from the previous diffusion model, we introduce an adaptation module $f_{adapt}$ based on a MLP. The diffusion model's output $z_1$, serves as the input to the adaptation module. This module aligns the distribution learned by the diffusion model with the physical constraints of hand-object interactions. The specific formula is as follows:
\begin{equation}
  \begin{aligned}
    z_2 = f_{adapt}(z_1)
  \end{aligned}
  \label{eqa:adapt_loss}
\end{equation}
where $z_2 \in R^{N_z}$ , is the output of the adaptation module when given $z_1$ as input. 

The goal of incorporating hand-object physical constraints is to ensure that the resulting hand mesh achieves natural and realistic grasping postures. However, $z_1$ and $z_2$ do not accurately represent the quality of hand-object interactions in real physical space. Therefore, we first reconstruct $z_1$  and $z_2$ back to the MANO parameters $h_p$,  and then use the MANO Layer\cite{mano} $f_{mano}$ to reconstruct the hand mesh $h_m$:
\begin{equation}
  \begin{aligned}
    h_p = \mathcal{D}(z_1 + z_2)
  \end{aligned}
  \label{eqa:Decoder}
\end{equation}
\begin{equation}
  \begin{aligned}
    h_m = f_{mano}(h_p)
  \end{aligned}
  \label{eqa:mano-layer}
\end{equation}
Next, we update the adaptation module using the loss function \ref{eq:Total-loss} to ensure that the physical constraints of hand-object interactions are accurately aligned. This training method addresses the challenge of directly learning physical constraints in diffusion models, resulting in more natural grasping poses and minimizing unnecessary physical penetration.


\subsection{Inference}
\label{inference}
Fig. \ref{fig:inference-pipeline} illustrates the inference process of our method. During inference, the initial input consists of noise $u$ sampled from a Gaussian distribution and an object point cloud $o_p$.

First, we generate the prior $z_1$ for the hand mesh in the latent space through an $N$-step denoising process~\cite{ddim}. Next, the adaptation module integrates $z_1$ with the object point cloud information to generate $z_2$, as shown in Eq. \ref{eqa:adapt_loss}. Finally, $z_1$ and $z_2$ are combined, and the decoder converts them into MANO parameters $h_p$, which are then used by the MANO layer \cite{mano} to produce the hand mesh $h_m$. This process can be described by the equations \ref{eqa:Decoder} and \ref{eqa:mano-layer}.

While using Diffusion Models (DDPM) for generating grasp postures marks a significant advancement over the previous two-stage model, there is still a need to enhance generation speed to meet practical requirements. To address this, we employ DDIM \cite{ddim}, which optimizes both speed and quality by adjusting the step size during the denoising process. This approach enables the rapid generation of grasping poses.


%% file: sec/4_exp.tex
\section{Experiment}
\begin{table*}[t]
    \centering
    \setlength{\abovecaptionskip}{0pt}
    \setlength{\belowcaptionskip}{0pt}
    \resizebox{\linewidth}{!}{
    \begin{tabular}{l|l|ccccc}
        \hline
        Dataset  & Details &
        \begin{tabular}{c} Penetration \\ Volume $\downarrow$ \end{tabular} &
        \begin{tabular}{c} Simulation \\ Displacement $\downarrow$ \end{tabular} &
        \begin{tabular}{c} Contact \\ Ratio $\uparrow$ \end{tabular} &
        \begin{tabular}{c} Entropy $\uparrow$ \end{tabular} &
        \begin{tabular}{c} Cluster \\ Size $\uparrow$ \end{tabular}  \\
        \hline
       OakInk~\cite{YangCVPR2022OakInk} 
            &Baseline CVAE model& 13.08 &  1.78 &  \textbf{98} &  2.81 &  1.12 \\
            &Original diffusion model& 18.34 &  \textbf{1.45} &  \textbf{98} &  2.91 &  \textbf{5.24} \\
            &Original diffusion model with physical loss& 6.31 & 3.77  & 71 & 2.85  & 1.58  \\
            &Our whole pipeline& \textbf{4.37} & \textbf{1.45} & 94 & \textbf{2.92} & 4.96\\
        \hline
        
        GRAB~\cite{grab} 
            &Baseline CVAE model& 12.33 &  1.94 &  98 &  2.62 &  0.87 \\
            &Original diffusion model& 15.46 & 1.80 & 96 & 2.87 & \textbf{3.06} \\
            &Original diffusion model with physical loss&8.43 & 5.24 & 50 & 2.84 & 1.26 \\
            &Our whole pipeline&\textbf{1.25} & \textbf{1.67} & \textbf{100} & \textbf{2.93} & 1.87 \\
        \hline
        HO-3D~\cite{ho3d} 
            &Baseline CVAE model & 23.17 &  3.12 &  \textbf{100} &  2.64 &  0.93 \\
            &Original diffusion model& 16.64 & 2.18 & 90 & 2.87 &  4.04 \\
            &Original diffusion model with physical loss& 12.73& 3.87&  62&  2.87 &  1.37    \\
            &Our whole pipeline& \textbf{5.23} & \textbf{2.14} & 98 & \textbf{2.88} & 3.97 \\
        \hline
    \end{tabular}
    }
    \caption{Ablation study results on the \textbf{GRAB, OakInk, HO-3D} datasets~\cite{grab,YangCVPR2022OakInk,ho3d}. 
    The evaluation of the HO-3D is an out-of-domain generalization test, where the model is trained using the GRAB dataset. 
    }
    \label{tab:ablation-total}
\end{table*}

\begin{figure*}[t]
\centering
    \setlength{\abovecaptionskip}{0pt}
    \setlength{\belowcaptionskip}{0pt}
    \begin{minipage}{0.02\textwidth} 
        \raggedleft
        \raisebox{0.8cm}{\rotatebox{90}{\small w/o AM}} 
        \raisebox{0.8cm}{\rotatebox{90}{\small AM}}
    \end{minipage}%
    \begin{minipage}{0.3\textwidth} 
        \centering
        \includegraphics[width=\textwidth]{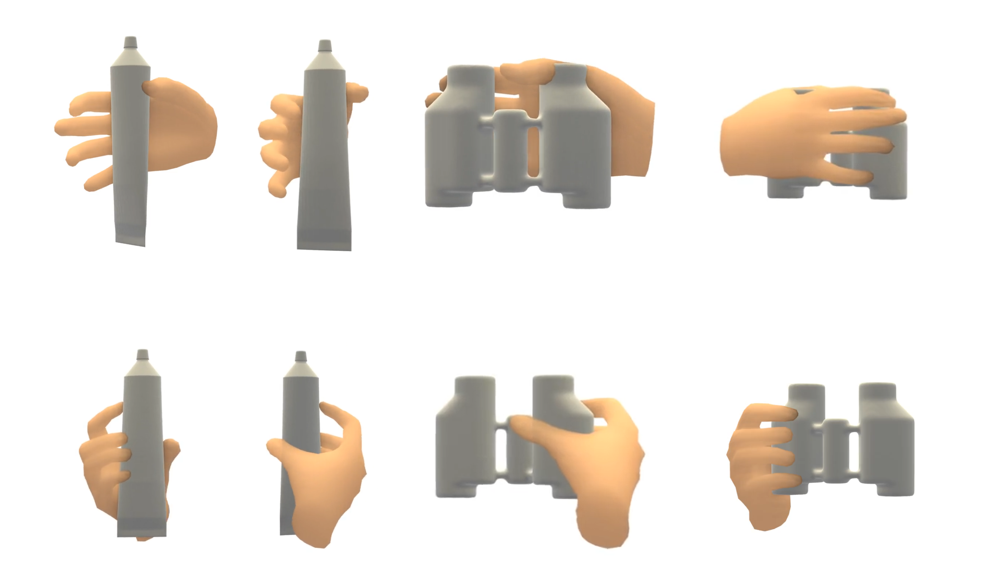}
    \end{minipage}%
    \vrule width 0.5pt 
    \begin{minipage}{0.3\textwidth} 
        \centering
        \includegraphics[width=\textwidth]{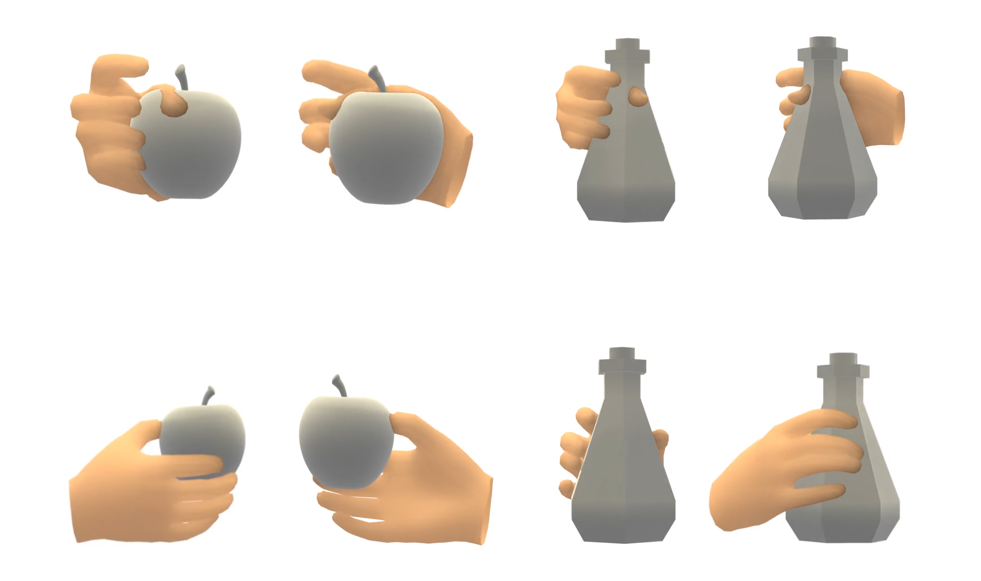}
    \end{minipage}%
    \vrule width 0.5pt 
    \begin{minipage}{0.3\textwidth} 
        \centering
        \includegraphics[width=\textwidth]{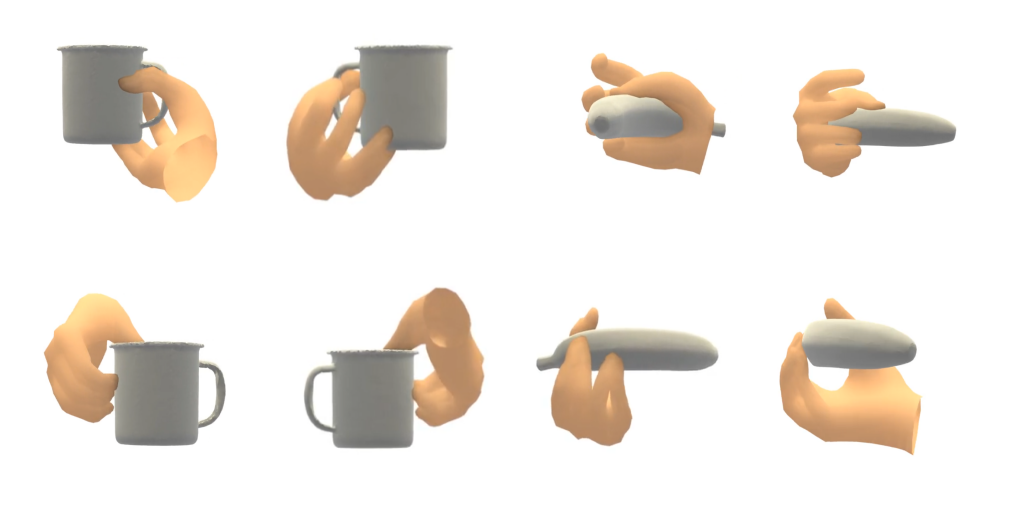}
    \end{minipage}

    \begin{minipage}{0.02\textwidth} 
        \raggedleft
        \raisebox{0.8cm}{\rotatebox{90}{\small w/o AM}} 
        \raisebox{0.8cm}{\rotatebox{90}{\small AM}}
    \end{minipage}%
    \begin{minipage}{0.3\textwidth} 
        \centering
        \includegraphics[width=\textwidth]{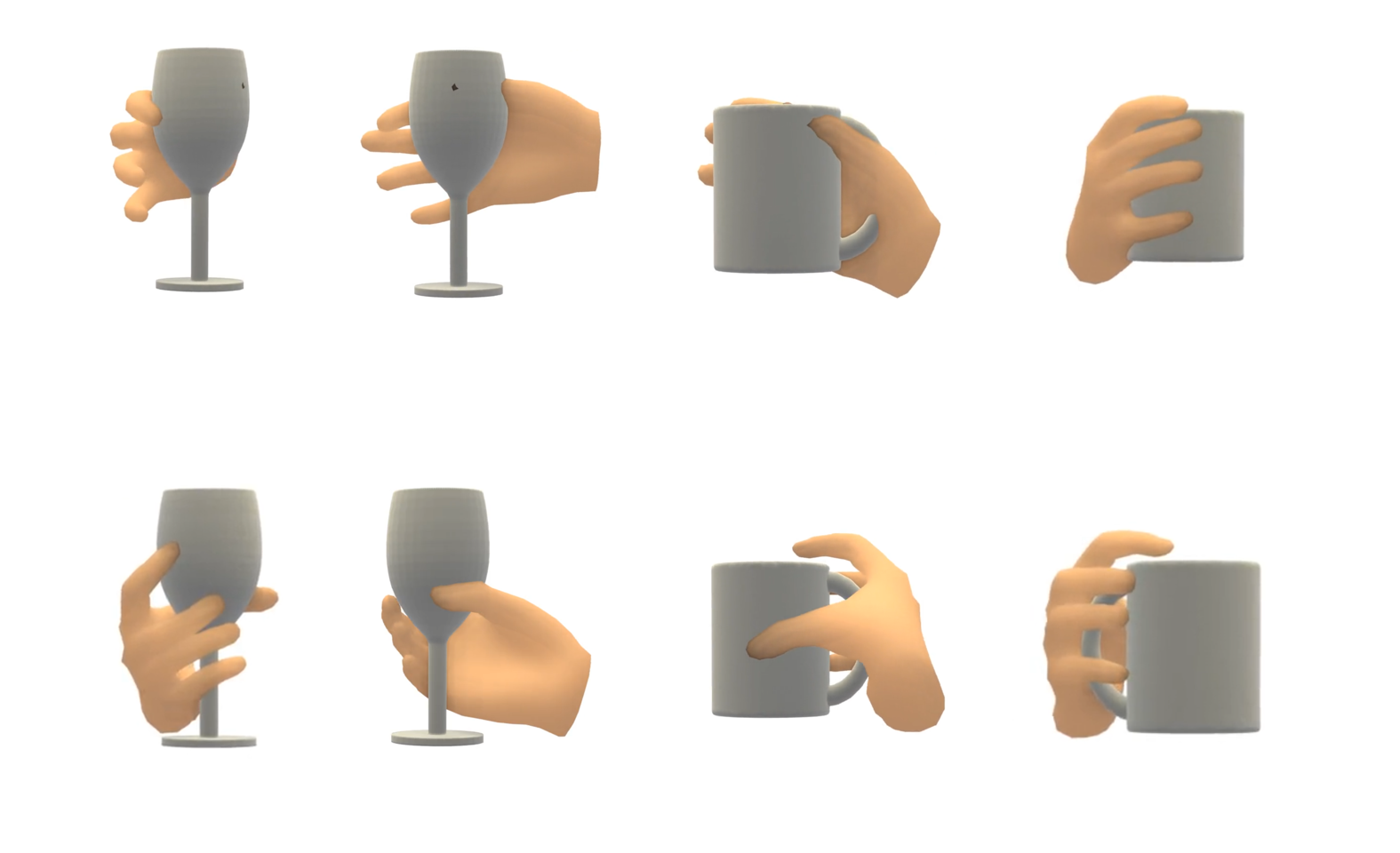}
        \captionsetup{labelformat=empty} 
        \caption*{GRAB~\cite{grab}}
        \label{fig:ablation-grab}
    \end{minipage}%
    \vrule width 0.5pt 
    \begin{minipage}{0.3\textwidth} 
        \centering
        \includegraphics[width=\textwidth]{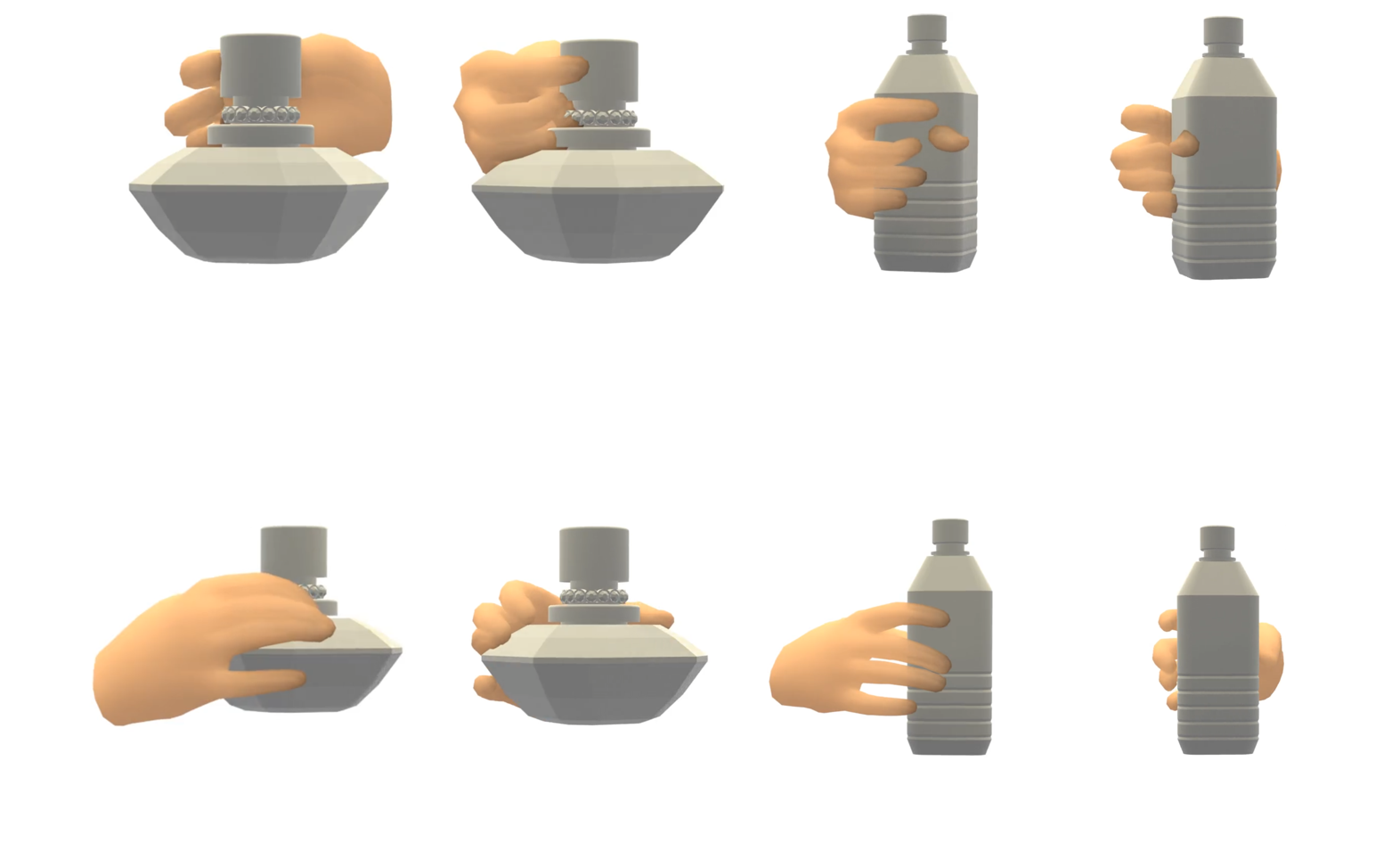}
        \captionsetup{labelformat=empty} 
        \caption*{OakInk~\cite{YangCVPR2022OakInk}}
        \label{fig:ablation-OakInk}
    \end{minipage}%
    \vrule width 0.5pt 
    \begin{minipage}{0.3\textwidth} 
        \centering
        \includegraphics[width=\textwidth]{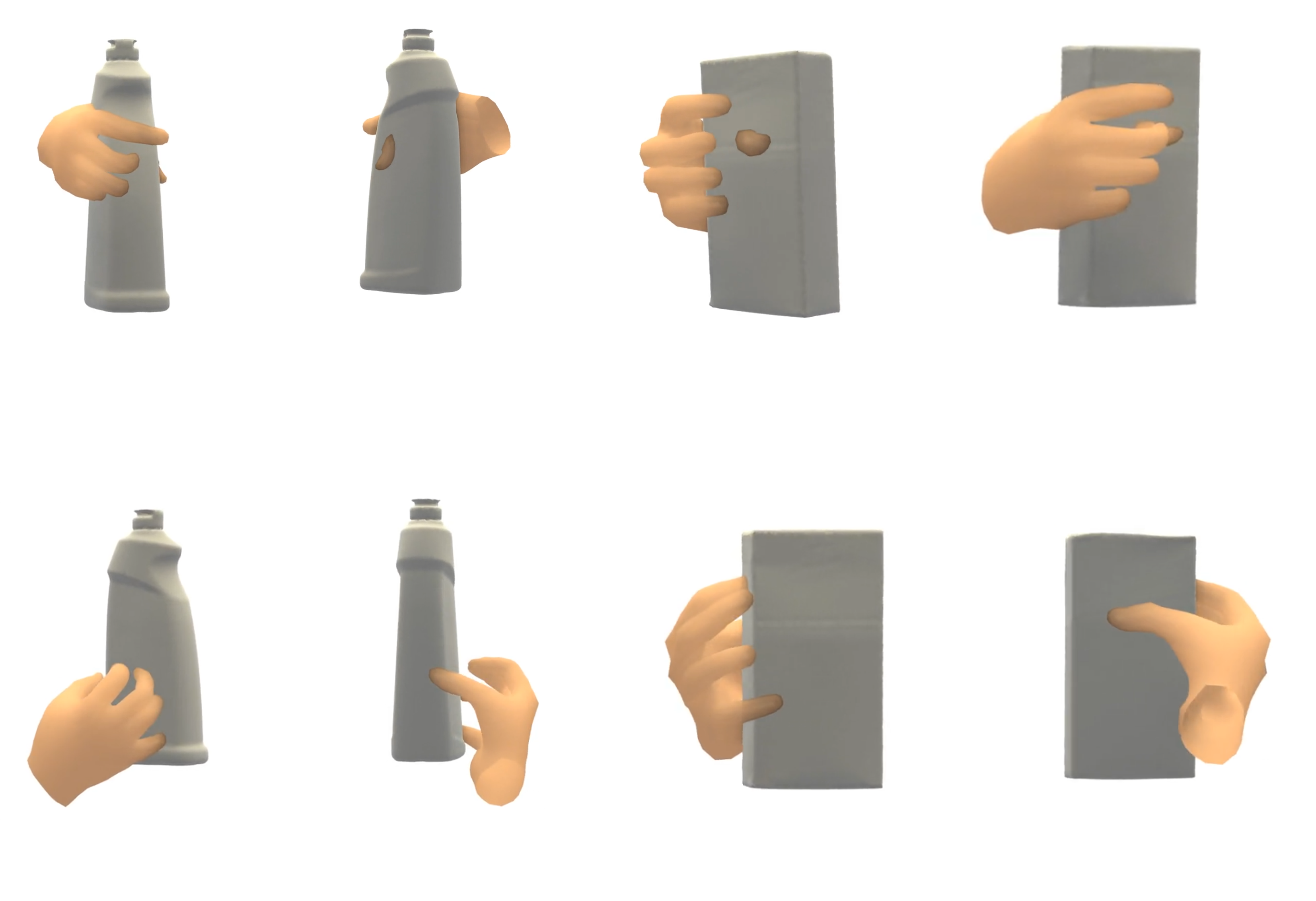}
        \captionsetup{labelformat=empty} 
        \caption*{HO-3D~\cite{ho3d}}
        \label{fig:ablation-ho3d}
    \end{minipage}
    \caption{Qualitative comparison between our method and Ours w/o Adaptation Module (AM). Starting from the same random Gaussian noise, we visualize the generated grasps by our whole pipeline (first row) and ours w/o Adaptation Module. For each object, we show two different views for visualization (two columns). This comparison demonstrates that our whole pipeline with AM notably reduces object penetration and produces more realistic grasp poses. }
    \label{fig:ablation-total}
\end{figure*}
In this section, we evaluate the effectiveness and efficiency of the proposed framework for object-conditioned hand pose generation. The structure is organized as follows. 

We first introduce our benchmarking datasets (Sec. \ref{sec:dataset}), evaluation metrics (Sec.~\ref{sec:evaluation_metrics}), and implementation details (Sec.~\ref{sec:imple_details}). 
Then, we conduct a model analysis to demonstrate the efficacy of each component in the proposed framework (Sec.~\ref{ablation}). 
In what follows, we compare our method with the recent state-of-the-art approaches (Sec.~\ref{experiment}). 
Finally, we assess the perceived quality and stability of the generated grasping poses through user studies (Sec. \ref{user-study}).


For experimental settings, we assess the model’s generalization to new objects using the out-of-domain dataset ~\cite{ho3d}. We also evaluate the physical penetration and grasp firmness of the generated poses with an in-domain setting on the OakInk and GRAB datasets~\cite{YangCVPR2022OakInk, grab}.
\subsection{Datasets}
\label{sec:dataset}
\begin{table*}[t]
    \centering
    \setlength{\abovecaptionskip}{0pt}
    \setlength{\belowcaptionskip}{0pt}
    \resizebox{\linewidth}{!}{
    \begin{tabular}{c|cc|cccc}
        \hline
        Method &
        \begin{tabular}{c} Penetration \\ Volume $\downarrow$ \end{tabular} &
        \begin{tabular}{c} Simulation \\ Displacement $\downarrow$ \end{tabular} &
        \begin{tabular}{c} Contact \\ Ratio $\uparrow$ \end{tabular} &
        \begin{tabular}{c} Entropy $\uparrow$ \end{tabular} &
        \begin{tabular}{c} Cluster \\ Size $\uparrow$ \end{tabular} &
        \begin{tabular}{c} Inference \\ Time $\downarrow$ \end{tabular} \\
        \hline
        GrabNet~\cite{grab} & 15.50 & 2.34 & 99 & 2.80 & 2.06 & 0.23s \\
        GraspTTA~\cite{tta} & 7.37 & 5.34 & 76 & 2.70 & 1.43 & 6.90s \\
        HALO~\cite{halo} & 25.84 & 3.02 & 97 & 2.81 & 4.87 & 10.42s \\
        GF~\cite{gf} & 93.01 & - & \textbf{100} & 2.75 & 3.44 & 32.75s \\
        ContactGen~\cite{liu2023contactgen} & 9.96 & 2.70 &  97 & 2.81 & \textbf{5.04} & 110.60s \\
       $ \text{Ours}^1$ & \textbf{5.23} & \textbf{2.14} & 98 & \textbf{2.88} & 3.97 & \textbf{0.14s} \\
        \hline
        $\text{ContactGen}^2$ & 14.32 & 2.41 &  \textbf{100} & 2.84 & \textbf{5.23} & 110.60s \\
        $ \text{Ours}^2$ & \textbf{12.30} & \textbf{1.44}  & \textbf{100}  & \textbf{2.88} & 4.41 & \textbf{0.14s} \\
        \hline
    \end{tabular}
    }
    \caption{Comparison with previous methods on the \textbf{HO-3D} dataset~\cite{ho3d}, where $\text{Ours}^1$ indicates our model is trained on the GRAB~\cite{grab} dataset following~\cite{gf,liu2023contactgen,halo}, and $\text{Ours}^2$ and $\text{ContactGen}^2$ suggests the corresponding models are trained on the OakInk~\cite{YangCVPR2022OakInk} dataset. Our model achieves state-of-the-art performance on this out-of-domain dataset, setting new benchmarks with faster inference speeds and the best physical metrics for generated grasps.}
    \label{tab:HO3D}
\end{table*}
We conduct experiments using the OakInk~\cite{YangCVPR2022OakInk}, GRAB~\cite{grab}, and HO-3D~\cite{ho3d} datasets, adhering to the experimental protocols outlined in~\cite{halo, liu2023contactgen, YangCVPR2022OakInk}. Specifically, in Sec. \ref{out-of-domain}, We train the model separately on the OakInk and GRAB datasets, and then evaluate its generalization ability on the HO-3D dataset. In Sec. \ref{in-domain}, we perform both training and evaluation on the OakInk and GRAB datasets.

The OakInk and GRAB datasets~\cite{YangCVPR2022OakInk, grab} consist of hand-object mesh pairs with hand models parameterized by the MANO~\cite{mano} model. The GRAB dataset includes real human grasps for 51 objects across 10 subjects, whereas the OakInk dataset features real human grasps for 1,700 objects from 12 subjects. Following~\cite{liu2023contactgen,grab,tta}, we also evaluate the model's generalization ability by testing on out-of-domain objects from the HO3D dataset.
\subsection{Evaluation Metrics}
\label{sec:evaluation_metrics}
Following the prior evaluation protocals~\cite{grab, gf, halo, liu2023contactgen, tta, wu2022saga}, we evaluate the generated grasping poses using the following criteria: (1) physical plausibility, (2) stability, (3) diversity, (4) generation speed, and (5) perception score.

\noindent\textbf{Physical Plausibility Assessment.} We evaluate physical plausibility by measuring hand-object mutual penetration volume and contact ratio~\cite{gf, halo, liu2023contactgen, tta}. The penetration volume is calculated by voxelizing the mesh into $1mm^3$ cubes and computing the overlapping voxels. The contact ratio indicates the proportion of the grasps in contact with the object.

\noindent\textbf{Grasp Stability Assessment.} Following~\cite{gf,wxf,wu2022saga,grab,liu2023contactgen,tta,YangCVPR2022OakInk}, we use a simulator to position the object and the generated grasps. We then measure the average displacement of the object's center of gravity due to gravity.

\noindent\textbf{Diversity Assessment.} We assess the diversity of generated grasps following~\cite{halo,liu2023contactgen}. First, we cluster the grasps into 20 clusters using K-means. Diversity is measured by computing the entropy of cluster assignments and the average cluster size, with higher entropy values and larger cluster sizes indicating greater diversity. Consistent with previous work, K-means clustering~\cite{halo, liu2023contactgen} is applied to 3D hand keypoints across all methods.

\noindent\textbf{Generation Speed Assessment.} We randomly select 128 objects from the dataset, generate grasping poses for each object, and calculate the average time required to generate a single pose on an NVIDIA A40 GPU.

\noindent\textbf{Perceptual Score Assessment.} We conduct a perceptual study, as described in~\cite{halo,tta}, with human participants to evaluate the naturalness of the generated grasps.
\subsection{Implementation Details}
\label{sec:imple_details}
During training, we use the Adam optimizer, LR = $1e^{-4}$,  $N_z = 768 , N_o = 3000$ and bath size = 256. During training the autoencoder, the loss weights are $\lambda_1 = 0.1, \lambda_2 = 1, \lambda_3 = 1000, \lambda_4 = 10, \lambda_5 =10 $. When training the diffusion model, we freeze the auto-encoder and sample 3000 points from the object mesh $o_m$ as the input point cloud $o_p$. When training the adaptation module, we use the same input point cloud and the loss weights are $\lambda_1^d = 100, \lambda_2^d = 0.1, \lambda_3^d = 1000, \lambda_4^d = 20, \lambda_5^d =0.1$.
\subsection{Ablation Study}
\label{ablation}
In this section, we conduct an ablation study to systematically evaluate the contribution of each module to the overall framework performance. This approach clarifies the role and impact of each component before delving into a detailed analysis of the experimental results.


Tab. \ref{tab:ablation-total} summarizes the results, showing that while the CVAE model slightly outperforms the diffusion model in penetration rate, it exhibits weaker generative performance, as indicated by lower entropy and smaller cluster sizes. Conversely, the diffusion model excels in entropy and cluster size but struggles with higher penetration, suggesting difficulties capturing the physical constraints of hand-object interactions. Integrating a physical loss function directly into the diffusion model decreases performance by increasing displacement and reducing grasp robustness, underscoring the challenge of aligning hand representations with physical constraints in latent space. Our Adaptation Module approach effectively combines the diffusion model with physical constraints, achieving reduced penetration and displacement, and significantly improving the accuracy of hand-object interactions.

Fig. \ref{fig:ablation-total} shows that our Adaptation Module method significantly enhances performance across all three datasets, reducing penetration volume and improving generalization on the out-of-domain HO-3D dataset. This improvement further demonstrates the Adaptation Module's ability to transform distributions, aligning the generated hand latent vector with natural human expectations.
\begin{table*}[t]
    \centering
    \setlength{\abovecaptionskip}{0pt}
    \setlength{\belowcaptionskip}{0pt}
    \resizebox{\linewidth}{!}{
    \begin{tabular}{c|c|cc|ccc}
        \hline
        Dataset & Method &
        \begin{tabular}{c} Penetration \\ Volume $\downarrow$ \end{tabular} &
        \begin{tabular}{c} Simulation \\ Displacement $\downarrow$ \end{tabular} &
        \begin{tabular}{c} Contact \\ Ratio $\uparrow$ \end{tabular} &
        \begin{tabular}{c} Entropy $\uparrow$ \end{tabular} &
        \begin{tabular}{c} Cluster \\ Size $\uparrow$ \end{tabular}  \\
        \hline
        OakInk~\cite{YangCVPR2022OakInk} & GrabNet~\cite{YangCVPR2022OakInk} & 6.60 & \textbf{1.21} & 94 & 1.68 & 1.22 \\
          & ContactGen* & 4.85 & 2.01 & 94 & 2.88 & 4.07 \\
          & Ours & \textbf{4.37} & 1.45 & \textbf{94} & \textbf{2.92} &\textbf{ 4.96} \\
        \hline
        GRAB~\cite{grab} & GrabNet~\cite{grab} & 1.72 & 3.65 & 96 & 2.72 & 1.93 \\
                         & HALO~\cite{halo} & 2.09 &  3.61 & 94 & 2.88 & 2.15 \\
                         & ContactGen~\cite{liu2023contactgen} & 2.16 &  2.72 & 96 & 2.88 & \textbf{4.11} \\
                         & Ours & \textbf{1.25} & \textbf{ 1.67} & \textbf{100} & \textbf{2.93} & 1.87 \\
        \hline
    \end{tabular}
    }
    \caption{Quantitative comparison on the \textbf{OakInk} and \textbf{GRAB} dataset~\cite{grab,YangCVPR2022OakInk}, where * indicates the model is trained on the OakInk dataset using the code released by the authors. 
    Our method achieves the best performance on almost all evaluation metrics.  
    }
    \label{tab:2}
\end{table*}
\begin{figure*}[t]
\centering
    \begin{minipage}{0.03\textwidth} 
        \raggedleft
        \raisebox{0.2cm}{\rotatebox{90}{\small GrabNet}} \\[0.1cm]
        \raisebox{0cm}{\rotatebox{90}{\small ContactGen}} \\[0.3cm]
        \raisebox{1cm}{\rotatebox{90}{\textbf{\small Ours}}}
    \end{minipage}%
    \begin{minipage}{0.3\textwidth} 
        \centering
        \includegraphics[width=\textwidth]{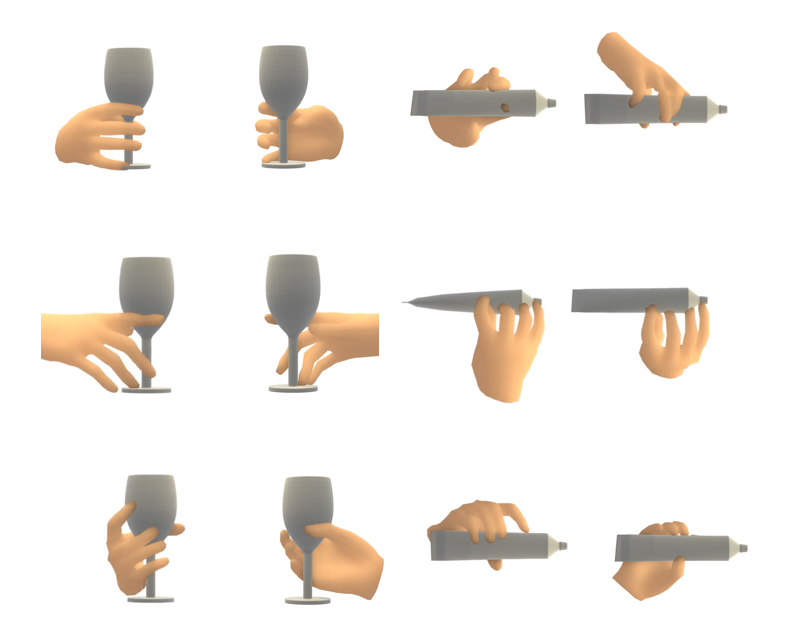}
        \captionsetup{labelformat=empty} 
        \caption*{GRAB~\cite{grab}}
        \label{fig:comparison-grab}
    \end{minipage}%
    \hspace{0.01\textwidth} 
    \vrule width 0.5pt 
    \hspace{0.01\textwidth} 
    \begin{minipage}{0.3\textwidth} 
        \centering
        \includegraphics[width=\textwidth]{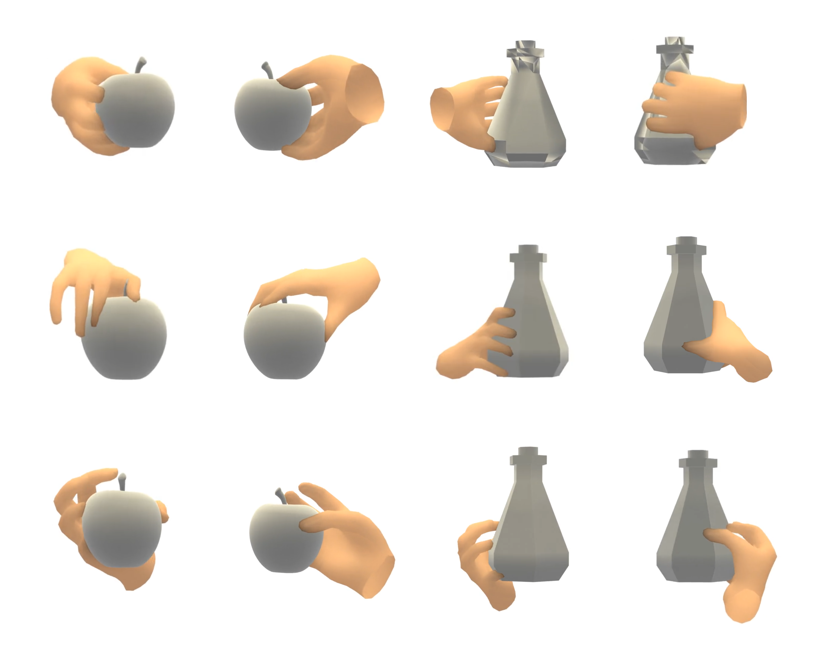}
        \captionsetup{labelformat=empty} 
        \caption*{OakInk~\cite{YangCVPR2022OakInk}}
        \label{fig:comparison-OakInk}
    \end{minipage}%
    \hspace{0.01\textwidth} 
    \vrule width 0.5pt 
    \hspace{0.01\textwidth} 
    \begin{minipage}{0.3\textwidth} 
        \centering
        \includegraphics[width=\textwidth]{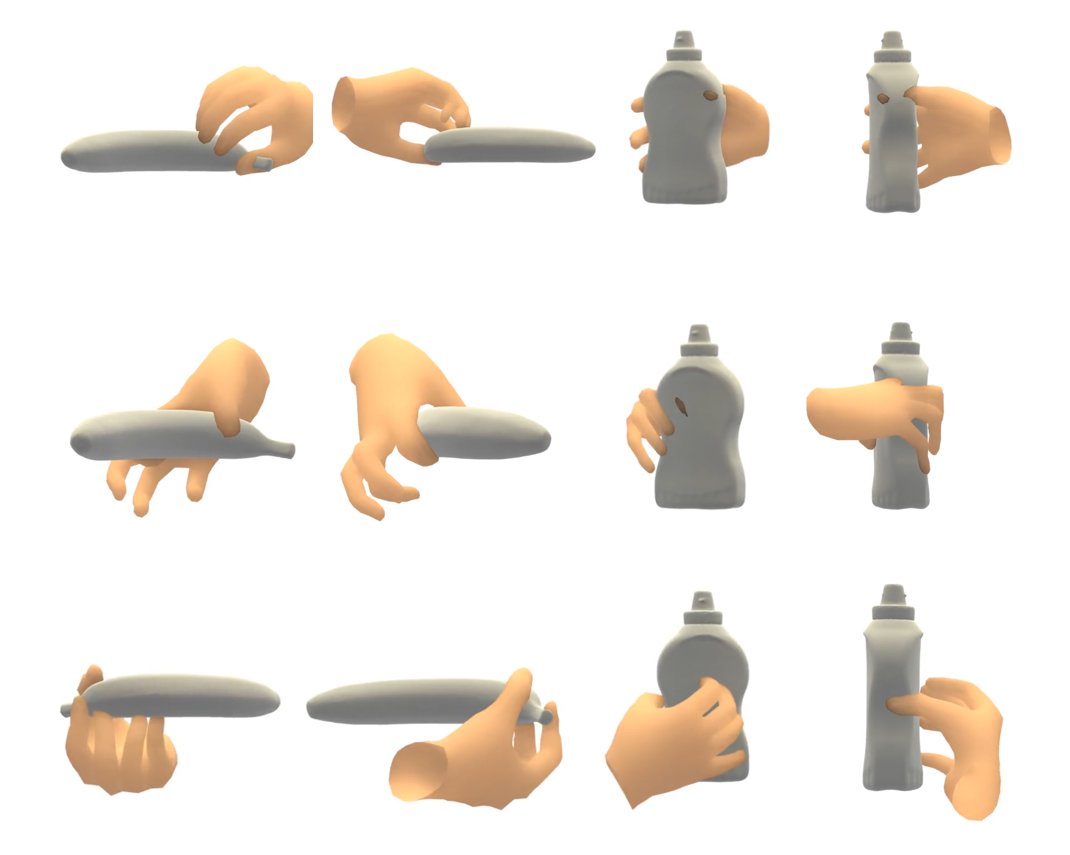}
        \captionsetup{labelformat=empty} 
        \caption*{HO-3D~\cite{ho3d}}
        \label{fig:comparison-ho3d}
    \end{minipage}
    \caption{Qualitative comparisons with state-of-the-art methods on GRAB, OakInk, and HO-3D datasets. Each pair (two columns) visualizes the generated grasps from two different views. Our method demonstrates a significant reduction in object penetration compared to other methods.}
    \label{fig:comparison}
\end{figure*}
\subsection{Grasp Generation Performance}
\label{experiment}

\noindent \textbf{Out-of-Domain.} We assess the generalization ability of our model using the HO-3D dataset~\cite{ho3d}. As demonstrated in Tab.~\ref{tab:HO3D} and Fig.~\ref{fig:comparison}, our method achieves the fastest generation speed, superior physical constraints, and entropy. In comparison, GrabNet~\cite{grab} matches our method in generation speed but suffers from significant physical penetration. ContactGen excels in cluster size but has the longest generation time, making it impractical for real-world applications. Overall, our method outperforms previous approaches in both physical generalization and generation speed.
\label{out-of-domain}
\noindent\textbf{In-Domain.} Tab. \ref{tab:2} and Fig. \ref{fig:comparison} compare our method with ContactGen~\cite{liu2023contactgen} and GrabNet~\cite{grab} on the OakInk dataset. Our method excels in penetration, contact ratio, entropy, and cluster size. Although displacement is slightly higher than GrabNet, our method achieves significantly lower penetration volume, demonstrating a better balance between minimizing physical intrusion and improving grasping effectiveness.\label{in-domain}

Tab. \ref{tab:2} compares our method with ContactGen~\cite{liu2023contactgen}, Halo~\cite{halo}, and GrabNet~\cite{grab} on the GRAB dataset. Our approach outperforms the others by achieving the lowest penetration and displacement and the highest contact ratio. Fig. \ref{fig:comparison} demonstrates that our method produces highly plausible object grasping. 
Although ContactGen produces more diverse grasps than our method in terms of cluster size, our method archives better results with smaller penetration and greater stability. 
By focusing on detailed geometric spatial information, our model creates more precise grasping poses. 
This precision increases entropy for objects with varied geometries, leading to more diverse hand poses, while similar object geometries result in more uniform grips and lower cluster sizes.
\begin{figure}[t]
    \centering
    \setlength{\abovecaptionskip}{0pt}
  \setlength{\belowcaptionskip}{0pt}
    \includegraphics[width=0.8\linewidth]{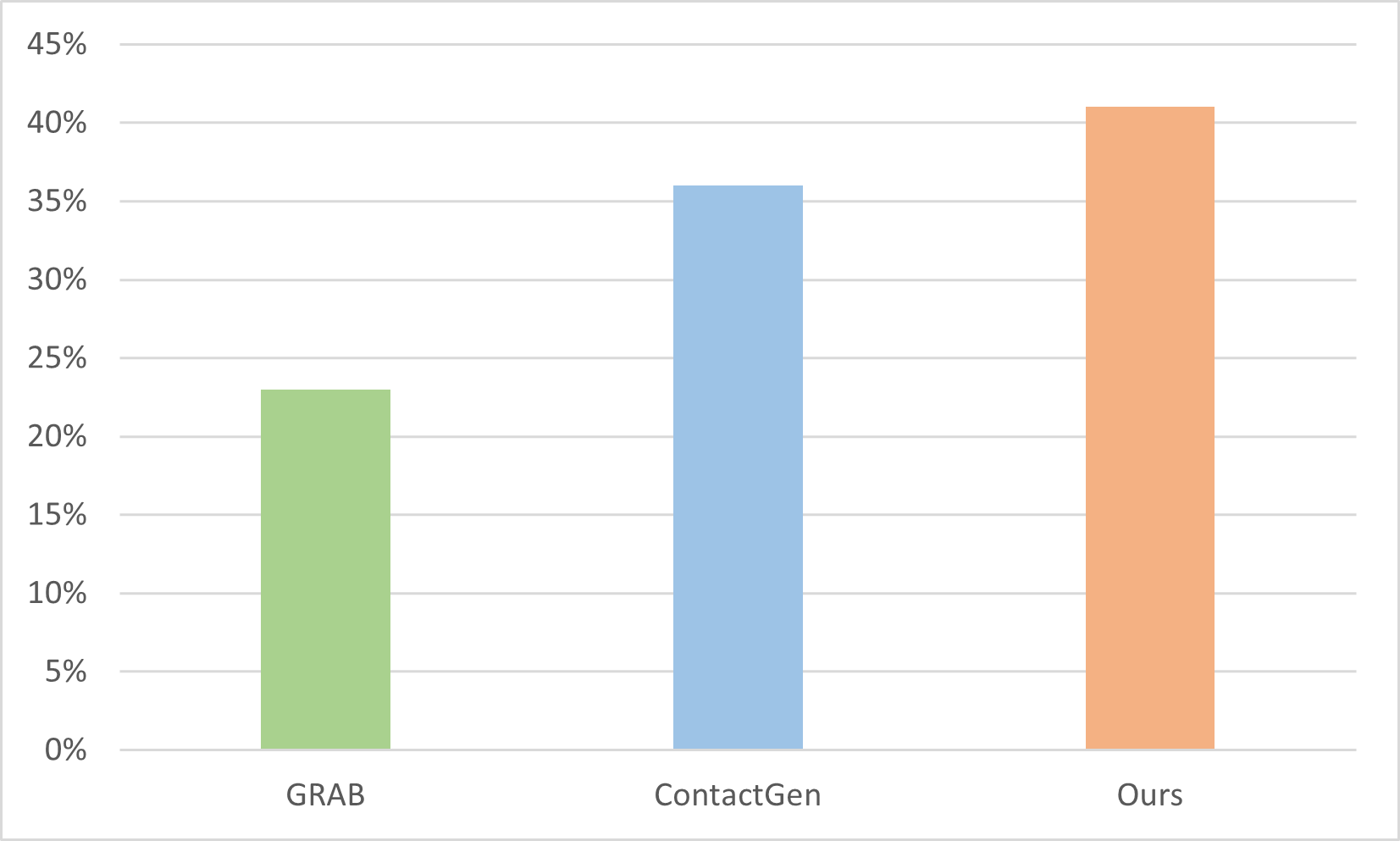}
    \caption{User study results. The numbers indicate the percentage of users who rate the corresponding method as more realistic.}
    \label{fig:user-study}
\end{figure}
\subsection{User Study}
\label{user-study}
We conduct a user study to evaluate the perceived quality and stability of grasps generated by different methods. We compare grasps generated by GrabNet~\cite{grab}, ContactGen~\cite{liu2023contactgen}, and our method by evaluating 10 objects from the GRAB~\cite{grab}, OakInk~\cite{YangCVPR2022OakInk}, and HO-3D~\cite{ho3d} datasets. Each object is tested with 3 grasps from each method. Ten participants select the best grasp pose based on the naturalness and stability of the grasp. Fig. \ref{fig:user-study} shows that our method received the highest number of selections in the experiment, indicating it generates the most natural and stable grasps.

%% file: sec/5_conclusion.tex
\section{Conclusion}
In this paper, we introduce a one-stage framework for rapid and realistic human grasp generation, eliminating the need for iterative optimization processes common in previous methods. We introduce an adaptation module that aligns the generative model’s output with physical constraints, refining hand representations in the latent space to enhance the accuracy and realism of generated grasps. Consequently, our method accelerates grasp generation, improves physical plausibility, and demonstrates robust generalization across diverse test inputs.

\section{Acknowledgement}
This work was supported by NSFC 62350610269, Shanghai Frontiers Science Center of Human-
centered Artificial Intelligence, and MoE Key Lab of Intelligent Perception and Human-Machine
Collaboration (ShanghaiTech University).

%% file: sec/X_suppl.tex
\clearpage
\twocolumn[{
\renewcommand\twocolumn[1][]{#1}%
\maketitlesupplementary
\begin{center}
    \vspace{-3ex}
    \captionsetup{type=figure}
    \includegraphics[width=\textwidth]{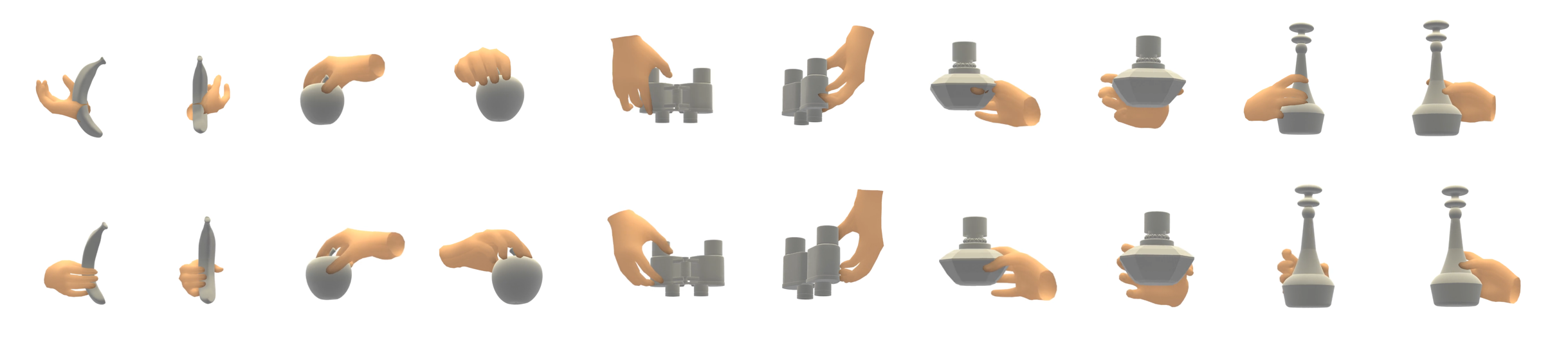}
    \vspace{-3ex}
    \caption{To assess the impact of a physically constrained loss function, we compare model performance with and without it. Each pair of columns shows generated grasps from two distinct views. The first row uses only the reconstruction loss, while the second row presents results from our proposed pipeline. Our method significantly reduces object penetration compared to using the reconstruction loss alone.}
    \label{fig:1}
     \captionsetup{type=figure}
    \includegraphics[width=\textwidth]{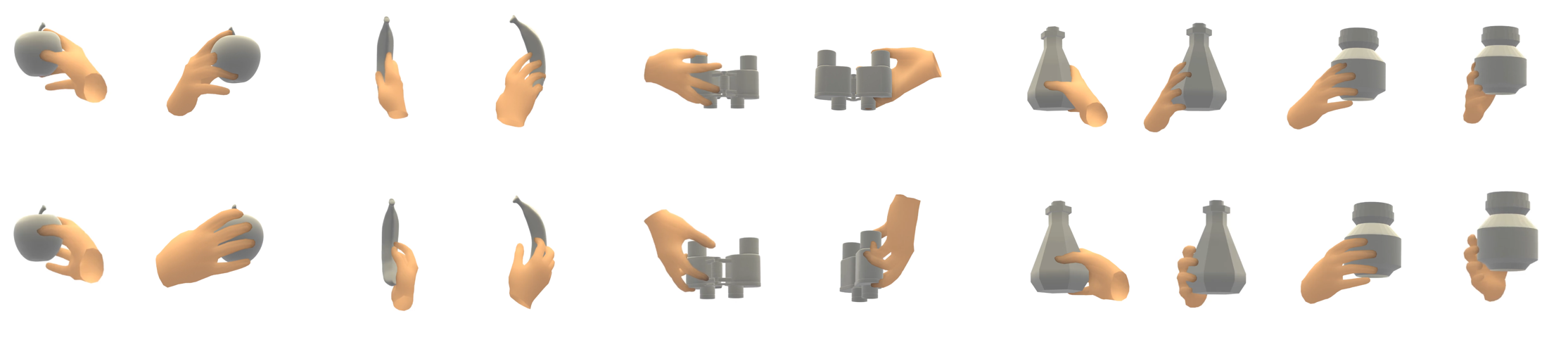}
    \vspace{-3ex}
    \caption{To evaluate the necessity of hand vertices as inputs, we visualize the model's output using both hand parameters and hand vertices. Each pair of columns shows generated grasps from two different views. The first row presents results with hand parameter input, while the second row displays results from our pipeline. Our method enhances performance by capturing hand joint details and improving rotational accuracy, which reduces object penetration.}
    \label{fig:2}
\end{center}
}]

\setcounter{page}{1}
\begin{table}[ht]
    \centering
    \resizebox{0.5\textwidth}{!}{ 
    \begin{tabular}{c|cccc}
        \hline
        OakInk &
        \begin{tabular}{c} Simulation \\ Displacement $\downarrow$ \end{tabular} &
        \begin{tabular}{c} Penetration \\ Distance $\downarrow$ \end{tabular} &
        \begin{tabular}{c} Penetration \\ Volume $\downarrow$ \end{tabular} &
        \begin{tabular}{c} Contact \\ Ratio $\uparrow$ \end{tabular} \\
        \hline
         No-physical-loss & 1.91 & 0.93 & 4.76 & 96 \\
         Hand param & \textbf{1.39} & \textbf{0.91} & 5.91 & \textbf{98} \\
         Ours & 1.83 & \textbf{0.91} & \textbf{2.39} & \textbf{98} \\
        \hline
    \end{tabular}
    }
    \caption{We conducted ablation experiments to evaluate the impact of the physical constraints loss function and hand vertices.}
    \label{tab:supplementart-table}
\end{table}

\section{Overview of Material}
The supplementary material comprehensively details our experiments, results, and visualizations. Tab. \ref{tab:supplementart-table} examines the impact of physical constraints during autoencoder training and compares the effects of hand verts versus hand parameters as inputs. Sec. \ref{vis} offers additional visualizations to enhance understanding of our model.
\section{More Autoencoder Experimental Results}
\label{experiment}
In training the autoencoder, we use hand vertices as input and apply both reconstruction and physical loss functions. Sec. \ref{2.1} and Sec. \ref{2.2} examine the effects of training the model with hand vertices and reconstruction loss alone versus using MANO parameters with both reconstruction and physical loss functions in Tab. \ref{tab:supplementart-table}.
\subsection{Training Using Reconstruction Loss} 
\label{2.1}
The model is trained using hand vertices $h_v$ as input and relies solely on the reconstruction loss function, without incorporating any physical loss function. As shown in Fig. \ref{fig:1}, experiments reveal that using only the reconstruction loss often results in significant penetration and displacement issues in hand-object interactions. However, as demonstrated in Tab.~\ref{tab:supplementart-table}, incorporating a physical constraint loss function improves the model's ability to capture these details, reducing physical collisions and enhancing grasp stability.
\subsection{Training Using Mano Parameter}
\label{2.2}
The model is trained using hand parameters $h_p$ as input. Our experiments indicate that using hand vertices instead of MANO parameters results in less physical volume intrusion. As shown in Fig. \ref{fig:2} and Tab. \ref{tab:supplementart-table}, this is attributed to the Hand vertices providing a more robust data representation than MANO parameters, reducing the model's sensitivity to input variations and thus improving training effectiveness.
\subsection{Autoencoder Visulization Result}
\label{vis}
\begin{figure*}[t]
\centering
\begin{minipage}{\textwidth}
    \centering
    \includegraphics[width=\textwidth]{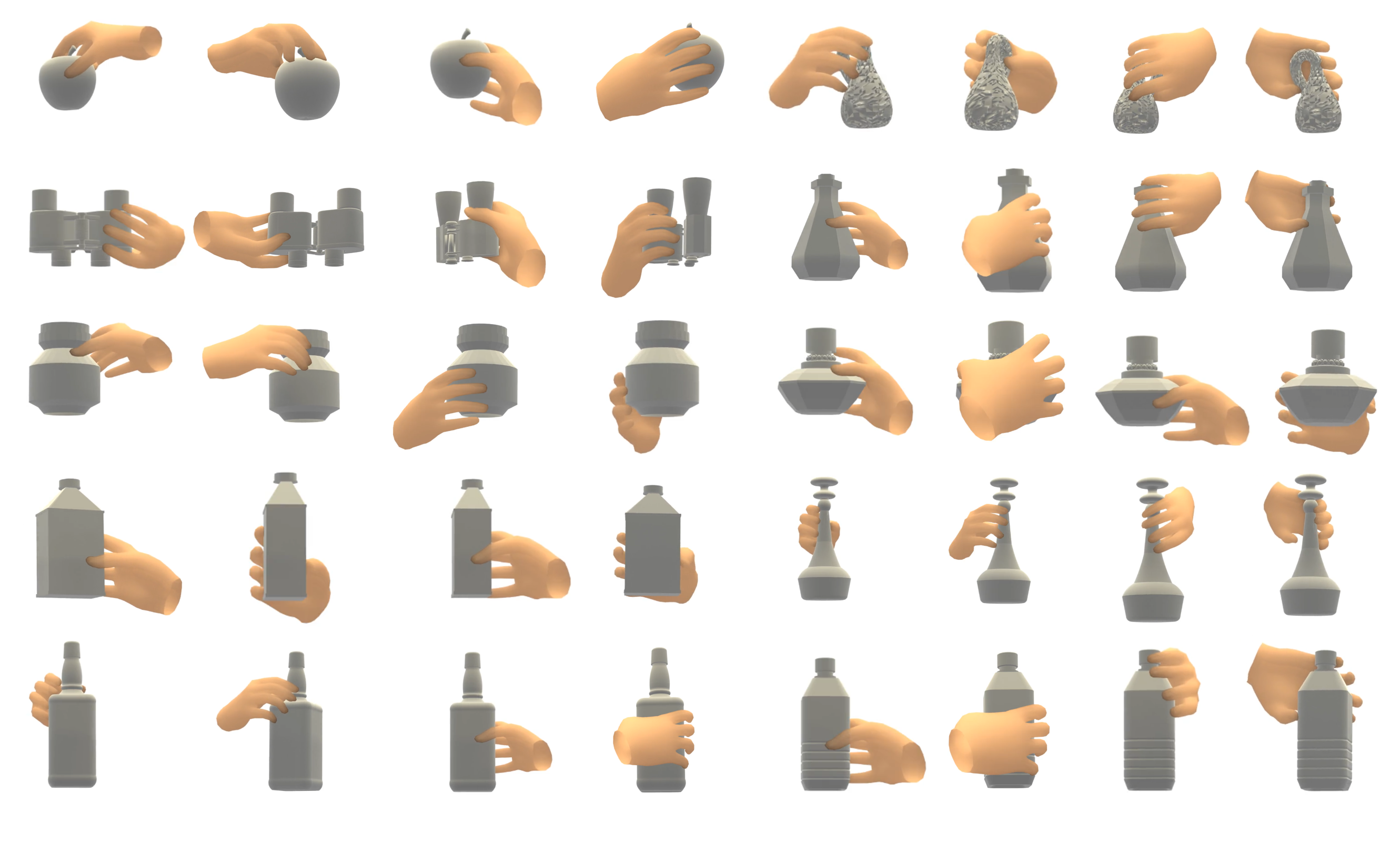}
    \caption{In the visualization results of the autoencoder, we selected two different grasping poses for each object, each shown from two different perspectives.} 
    \label{fig:ae-total-1}
\end{minipage}\hfill
\begin{minipage}{\textwidth}
    \centering
    \includegraphics[width=\textwidth]{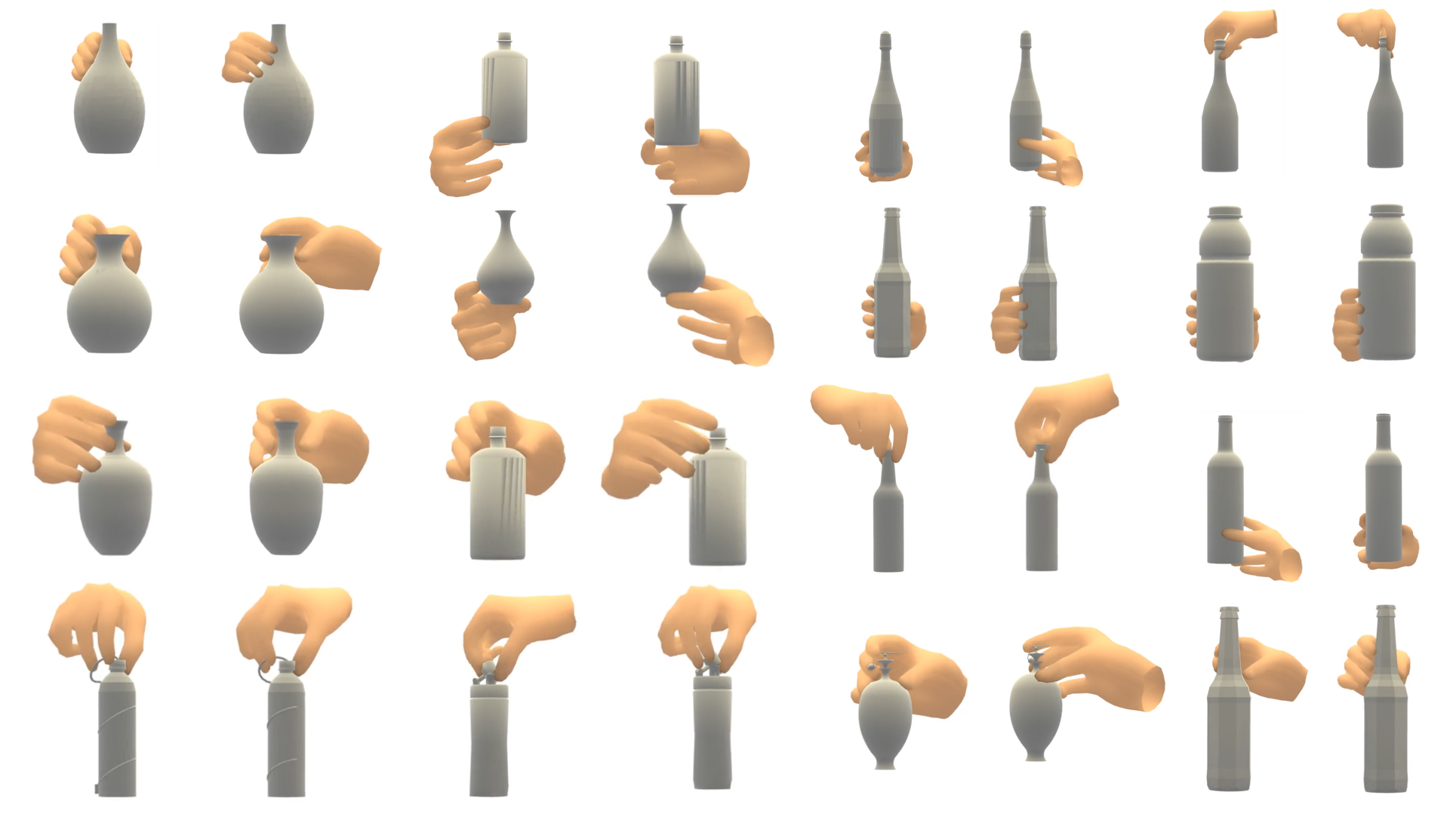}
    \caption{In the autoencoder visualization results, we randomly selected grasping poses, each shown from two different perspectives.}
    \label{fig:ae-total-2}
\end{minipage}
\end{figure*}

To validate the effectiveness of our autoencoder model, we provide extensive visualizations in Fig. \ref{fig:ae-total-1} and \ref{fig:ae-total-2}. 

Fig. \ref{fig:ae-total-1} illustrates two grasping poses for randomly selected test objects. This demonstrates that our model adheres to physical constraints in hand-object interactions for various grasps of the same object. Fig. \ref{fig:ae-total-2} showcases grasping poses for objects with diverse geometric shapes from the test set, highlighting our model's ability to generate effective grasps across different objects consistently.

